# A density compensation-based path computing model for measuring semantic similarity

Xinhua Zhu*, Fei Li, Hongchao Chen, Qi Peng
Guangxi Key Lab of Multi-source Information Mining & Security, Guangxi Normal University, Guilin 541004, China

Abstract

The shortest path between two concepts in a taxonomic ontology is commonly used to represent the semantic distance between concepts in the edge-based semantic similarity measures. In the past, the edge counting is considered to be the default method for the path computation, which is simple, intuitive and has low computational complexity. However, a large lexical taxonomy of such as WordNet has the irregular densities of links between concepts due to its broad domain but. The edge counting-based path computation is powerless for this non-uniformity problem. In this paper, we advocate that the path computation is able to be separated from the edge-based similarity measures and form various general computing models. Therefore, in order to solve the problem of non-uniformity of concept density in a large taxonomic ontology, we propose a new path computing model based on the compensation of local area density of concepts, which is equal to the number of direct hyponyms of the subsumers of concepts in their shortest path. This path model considers the local area density of concepts as an extension of the edge-based path and converts the local area density divided by their depth into the compensation for edge-based path with an adjustable parameter, which idea has been proven to be consistent with the information theory. This model is a general path computing model and can be applied in various edge-based similarity algorithms. The experiment results show that the proposed path model improves the average correlation between edge-based measures with human judgments on Miller and Charles benchmark from less than 0.8 to more than 0.85, and has a big advantage in efficiency than information content (IC) computation in a dynamic ontology, thereby successfully solving the non-uniformity problem of taxonomic ontology.

## 1 Introduction

The measurement of semantic similarity between concepts or words is an important fundamental research topic in natural language processing. It can be widely applied in the fields of intelligent retrieval [1], word sense disambiguation [2], machine learning [3], word spelling error detection and correction [4], machine translation [5] and text segmentation [6], and so on. In the late eighties and early nineties of last century, semantic similarity have been proven to be useful in some specific applications of computational intelligence and a number of semantic similarity methods have been proposed and developed [1,4,6,7,8,9], these methods can be divided into two groups [18]: One is edge counting-based methods, which uses the minimum number of edges of linking the corresponding ontological nodes for measuring similarity [11], and they are applied in some specific applications of computational intelligence with highly constrained taxonomies, such as medical semantic nets. Another group is information theory-based methods, which are based on a large-scale statistic corpus and utilizes the concept's information content (IC) that depends on the probability of encountering an instance of the concept in a corpus to measure semantic similarity between concepts [12]. These methods can be adapted to a particular application that has an approximate area with the corpus.

---

*Corresponding author
Email addresses: zxh429@263.net

With the emergence and development of large online semantic dictionary of WordNet [13], the researches on semantic similarity have turned to the more general applications, such as information extraction [14] and semantic annotation [15]. Through the continuous efforts of researchers, the methods of semantic similarity measures have been constantly improved and many new approaches are emerging, for example, the depth of concepts have been introduced into the edge counting-based methods [16], and a comprehensive IC intrinsic measurement approach has been proposed [27], which is connected only to the hierarchical structure of an ontology. Moreover, the feature-based measures [34] and hybrid approaches [35] have been proposed in succession.

At present, the edge-based and information content-based approaches are still the research focus of semantic similarity. Edge is an important component of the hierarchical structure of a taxonomic ontology, so the edge-based semantic similarity metric is intuitive, easy to understand and has low computational complexity. However, a large lexical taxonomy may has the irregular densities of links between concepts due to its broad domain [18], which would cause the same concept paths in the different density areas of a taxonomic ontology represent different semantic distance. At present this problem can't be effectively solved in the edge-based approaches, the excellent edge-based similarity approaches only get an about 0.8 correlation degree with human judgments in Miller & Charles (MC30) benchmark [4, 11, 16, 18, 19, 20, 21]. Though, the non-uniformity problem of taxonomic ontology can be better corrected by the information content-based approaches combined with the depth of concept in taxonomy hierarchy, they can get an about 0.85 correlation degree with human judgments in MC30 benchmark [17, 22]. But information content computation require to count the number of all hyponyms of the concept in the taxonomy [17,22], this is a complex computing process in a large taxonomic ontology, so information content-based similarity metric has a drawback of high computational complexity, which may prevent the popularization and application of this approach in a dynamic ontology. Now most of IC calculations assume that the number of hyponyms of each concept is well-known a priori and store them in the hash table of a file enabling an immediate computation in each measurement. However, in the big data era of rapid updating of information, the development trend of taxonomic ontology is online and real-time updates such as the Dbpedia Knowledge Base[37] based on Wikipedia, which represents real community agreement and automatically evolves as Wikipedia changes; as a result, the assumption of priori taxonomic ontology would not be true. Therefore, it is very important and urgent to find out a similarity approach that is equivalent to the performance of IC-based similarity metric and has lower computational complexity.

In this paper, we advocate that the path computation is able to be separated from the edge-based similarity measures to form various general computing models, as the information content computation can be isolated from the information theory-based similarity measures; and we exploit the local area density of concepts to establish a new path computing model based on the edge counting, which aim is to better correct the uneven density distribution in WordNet with less computational overhead. This model is a general path computing model and can be applied in most of edge-based similarity algorithms. The experiment results show that our path model can greatly improve the measurement accuracy of various edge-based similarity algorithms and has a huge advantage in computational complexity than the information content computation.

The rest of the paper is organized as follows: Section 2 provides an overview about the popular similarity approaches related to our study. Section 3 reorganizes the two existing edge-based path computing models and initially explores the approach for solving the non-uniformity problem of density of concepts in the taxonomic ontology. Section 4 proposes a new path computing model based on the local area density of concepts according to a widespread phenomenon in similarity measures and uses the information theory to prove its feasibility. Section 5 evaluates our path computing model from two aspects of performance and efficiency. Section 6 gives several conclusions according to the experiment results.

# 2 Related approaches

Currently the semantic dictionaries WordNet [13], VerbNet [23], FrameNet [24] and MindNet [25] can be used as the taxonomic ontology for similarity measures, and most of the popular similarity approaches are implemented and evaluated by using WordNet as the underlying reference ontology because of its clear concept hierarchy and abundant vocabulary. Here we introduce some popular similarity approaches related to our study.

## 2.1 Path-based approaches

The shortest path distance between concepts has a close relationship with their similarity. Actually, shortest concept path and concept similarity are the different forms of the same characteristics between a pair of concepts, in which a simple corresponding relationship can be created.

Rada et al. [11] exploited the shortest path length connecting two concepts via is-a links to measure their similarity. They defined the similarity between the concepts $c_1$ and $c_2$ as Eq. (1):

$$sim_{\text{Rada}}(c_1, c_2) = 2 \times MAX - P \quad (1)$$

Where $MAX$ is the maximum path length between concepts in the hierarchy of the taxonomic ontology, and $P$ is the shortest path length between the concepts $c_1$ and $c_2$, which is equal to the number of "is a" links from $c_1$ and $c_2$.

Leacock and Chodorow [19] mapped the shortest path length between two concepts into a similarity score using a logarithmic function. The similarity between the concepts $c_1$ and $c_2$ was defined as Eq. (2):

$$sim_{LC}(c_1, c_2) = -\log \frac{length(c_1, c_2)}{2D} \quad (2)$$

Where $D$ is the depth of the taxonomy hierarchy, it is used to scale the shortest path length.

Hirst and St-Onge [20] defined the similarity between concepts according to their path distance as Eq. (3):

$$sim_{HS}(c_1, c_2) = C - d - k * n \quad (3)$$

Where $d$ is the shortest path length between the concepts $c_1$ and $c_2$, and $n$ is the number of changes of direction in the path. $C$ and $k$ are the constant parameters, respectively let $C$ equal to 8 and $k$ equal to 1. When the path between the concepts $c_1$ and $c_2$ does not exist, which means that there is no similarity between them and their similarity is given as 0. However, this method is mainly used in the measurement of the two concepts relatedness.

## 2.2 Path and depth-based approaches

Wu and Palmer [16], in their study on lexical selection problems in machine translation,

proposed a new Path and depth-based approach on measuring semantic similarity. They defined the similarity between two concepts $c_1$ and $c_2$ as Eq. (4):

$$sim_{WP}(c_1,c_2)=\frac{2d}{2d+p_1+p_2} \quad (4)$$

Where $p_1$ and $p_2$ are respectively to represent the path length from the concept $c_1$ or $c_2$ to their least common subsumer, $d$ represents the path length from their least common subsumer to the root node, which is equal to the number of "is a" links from the least common subsumer to the root node.

Liu et al. [4] proposed a different method to measure concept semantic similarity based on path and depth. Their fundamental idea is to simulate the process of human judgment, which is based on the ratio of common features and different features between two concepts in the taxonomy hierarchy. They presented the following two equations:

$$sim_{Liu-1}(c_1,c_2)=\frac{\alpha\times d}{\alpha\times d+\beta\times p} \quad (5)$$

$$sim_{Liu-2}(c_1,c_2)=\frac{e^{\alpha\times d}-1}{e^{\alpha\times d}+e^{\beta\times p}-2} \quad (6)$$

Where $\alpha$ and $\beta$ are the smoothing factors for depth and path ($0<\alpha, \beta<1$), the $p$ is the shortest path length between the concepts $c_1$ and $c_2$, and the $d$ is the depth of their least common subsumer in the taxonomy hierarchy. The experiments showed that when the parameters $\alpha$ and $\beta$ were respectively taken 0.5, 0.55 in Eq. (5) and 0.25, 0.25 in Eq. (6), the measured similarity scores were closest to human judgments.

Li and Mclean [18] overcame the weakness of relying alone on the shortest path length between two concepts, and proposed a non-linear function to measure semantic similarity between two concepts. The proposed function is as Eq. (7):

$$sim_{LM}(c_1,c_2)=e^{-\alpha p}\frac{e^{\beta d}-e^{-\beta d}}{e^{\beta d}+e^{-\beta d}} \quad (7)$$

Where $\alpha$ and $\beta$ are the smoothing factors, its purpose is to scale the contribution of the $p$ and $d$ ($0<\alpha, \beta<1$). The $p$ is the shortest path length between the concepts $c_1$ and $c_2$, and the $d$ is the depth of their least common subsumer in the taxonomy hierarchy.

Hao et al. [21] attempted to imitate the thought process of human, and proposed a new method to compute the similarity between concepts. They defined the similarity formula as Eq. (8):

$$sim_{Hao}(c_1,c_2)=(1-\frac{p}{p+d+\beta})*(\frac{d}{p+d/2+\alpha}) \quad (8)$$

Where α and β are the smoothing factors, the $p$ is the shortest path length between the concepts $c_1$ and $c_2$, and the $d$ is the depth of their least common subsumer in the taxonomy hierarchy. The experiments showed that when $\alpha=0$, $\beta=1.0$, the measured similarity scores were closest to human judgments.

Sussna [26] proposed an edge weight-based method for measuring semantic similarity. He first computes the minimum distance between all adjacent nodes in the shortest path between the concepts $c_1$ and $c_2$, and then the semantic distance between the concept $c_1$ and $c_2$ is equal to the sum of these minimum distances. The distance formula is as Eq. (9):

$$\mathrm{dis}(c_1,c_2)=\sum MinDist(x,y) \mid \forall x,y \text{ in path}(c_1,c_2) \quad (9)$$

Where $x$ and $y$ are the adjacent nodes in the shortest path between the concept $c_1$ and $c_2$, MinDist(x, y) is the minimum distance between two adjacent nodes x and y. In the MinDist(x, y), the minimum distance isn't simply representing the number of edge between concepts, but each edge is assigned a weight, then Susssna exploits the depth-relative scaling of weight to calculate the MinDist(x, y). The minimum distance is defined as Eq. (10):

$$MinDist(x,y)=\frac{w(x\rightarrow_{r} y)+w(x\rightarrow_{r'} y)}{2\max[depth(x),depth(y)]} \tag{10}$$

Where $\rightarrow r$ is a relation of type *r* (synonym, hypernymy, hyponymy, holonymy, meronymy or antonymy), the $\rightarrow r'$ is the inverse relation of type *r*, for example, the type *r* represent hypernymy, and the type r' represent hyponymy. The weight of two adjacent nodes x and y is dedined as Eq. (11):

$$w(x\rightarrow_{r} y)=\max_{r}-\frac{\max_{r}-\min_{r}}{n_{r}(x)} \tag{11}$$

Where $n_r(x)$ is the number of relations of type r, leaving x. When the type r is hypernymy, hyponymy, holonymy or meronymy relation, the $\min_{r}$=1 and $\max_{r}$=2. When the type r is synonym and antonymy relations, the weight is equal to 0 and 2.5, respectively.

## 2.3 Information content-based approaches

### 2.3.1 Similarity measures based on IC

Resnik [12] was the first person to combine ontology and corpus, he stated that concept similarity depends on the amount of shared information between them, and propose an information content-based method. The similarity formula is as Eq. (12):

$$sim_{\mathrm{Res}}(c_1,c_2)=IC(LCS(c_1,c_2)) \tag{12}$$

Jiang and Conrath [22] proposed a distance-based method for measuring semantic similarity between concepts. The length of the taxonomical links is quantified as the difference between the IC of a concept and its subsumer. When computing the semantic distance between two concepts, they use the sum of the IC of each individual concept to subtract the IC of their LCS. The distance formula is as Eq. (13):

$$dis(c_1,c_2)=(IC(c_1)+IC(c_2))-2IC(LCS(c_1,c_2)) \tag{13}$$

Lin [17] proposed a method to measure semantic similarity based on information content(IC). He exploited the ratio of the commonalities between the concept $c_1$ and $c_2$ and their fully information-needed as the similarity score between concepts. The similarity formula is as Eq. (14):

$$sim_{Lin}(c_1,c_2)=\frac{2IC(LCS(c_1,c_2))}{IC(c_1)+IC(c_2)} \tag{14}$$

### 2.3.2 IC calculation methods

There are two main IC calculation models: corpora-based IC calculation and intrinsic IC calculation. The corpora-based IC calculation method was first proposed by Resnik [12], which requires a large corpus to count the probability of a concept, and was mainly used in the early stages. The intrinsic IC calculation method was first proposed by Seco [27], which is connected only to the hierarchical structure of a taxonomic ontology.

Resnik [12] was the first to propose the IC calculation method, and use the probability of concept *c* in a given environment. The IC value is as Eq. (15):

$$IC(c) = -\log(p(c)) \tag{15}$$

In the above Eq. (15), the p(c) is calculated as Eq. (16):

$$p(c) = \frac{\sum_{w \in Word(c)} count(w)}{N} \tag{16}$$

Where Word(c) is the set of words subsumed by the concept c, count(w) is the frequency of the word *w in* the corpus, and N is the total number of observed words in the corpus.

Seco et al. [27] proposed an intrinsic method for calculating IC, which only depend on the number of concept hyponyms in a taxonomic ontology. The IC of the concept *c* is calculated as Eq. (17):

$$IC(c) = 1 - \frac{\log(hypo(c)+1)}{\log(\max_nodes)} \tag{17}$$

Where $hypo(c)$ is the hyponym number of concept *c*, $\max_nodes$ is a constant (in WordNet3.0, $\max_nodes$ =82115).

Sánchez et al. [28] analyzed and discussed some of the better semantic evidence modelled, and proposed a new intrinsic IC computation model, which consider only the leaves of the concept's hyponym nets as an indication of its IC. Their IC computation model is as Eq. (18):

$$IC(c) = -\log\left(\frac{\frac{|leaves(c)|}{|subsumers(c)|}+1}{\max_leaves+1}\right) \tag{18}$$

The *leaves* and *subsumers* are defined as follows:

$$leaves(c) = \{l \in C \mid l \in hyponyms(c) \wedge l \text{ is a leaf}\} \tag{19}$$

$$subsumers(c) = \{a \in C \mid c \leq a\} \cup \{c\} \tag{20}$$

Where C is the set of concepts of the taxonomic ontology, c≤a means that *c* is *a* hierarchical specialization of *a*.

# 3. Path computing model

As information content computation can be isolated from the information theory-based similarity measures, we think the path computation can also be separated from the edge-based semantic similarity measures to form various general computing models. In order to obtain more accurate results in the edge-based measures, the way in which concepts' path is computed is crucial. Edge counting is the most common approach for computing concept path. This path computing model is simple, easy to understand and has low computational complexity, but is powerless for the non-uniformity of concept density in the taxonomic ontology. In addition, for trying to solve the non-uniformity problem of taxonomic ontology, we extract an edge weight-based path computing model from the edge-based similarity approach proposed by Sussna. Similarity measures mainly depend on the "is-a" relationship in a taxonomic ontology [29], we consider only the path between concepts in "is-a" relationship in this paper.

## 3.1 Edge counting-based path computing model

At present, most of edge-based semantic similarity measures [4,11,16,18,29,20,21] directly count the number of edges linking two concepts to calculate the length of shortest path between concepts. Assuming that the set of edges in shortest path between the concepts $c_1$, $c_2$ are represented by a function Edges(path($c_1$, $c_2$)), we can summarize the path computing model into the following formula (21):

$$Path(c_1, c_2) = |Edges(path(c_1, c_2))| \quad (21)$$

Where, path ($c_1$, $c_2$) refer to the shortest path between the concepts $c_1$, $c_2$ and | Edges (path($c_1$, $c_2$))| to the number of edges in the path ($c_1$,$c_2$).

## 3.2 Edge weight-based path computing model

In this study, we try to establish a density-based weight for edges according to the similarity approach proposed by Sussna et al. and form a general path computing model, which purpose is to explore the ways to solve the non-uniformity problem of taxonomic ontology by edge-based similarity approaches.

Assuming that path ($c_1$, $c_2$) refer to the shortest path between concepts $c_1$ and $c_2$ and it contains *n* edges, then the Path($c_1$,$c_2$) is computed as follows:

$$Path(c_1, c_2) = \sum_{i=1}^{n} Weight(e_i) \quad (22)$$

Where, Weight ($e_i$) is the weight of edge *e* in the path. Assuming that x, y are the two adjacent nodes linked by the edge *e*, Weight ($e_i$) is determined by the density of nodes x, y in taxonomic ontology as follows:

$$Weight(e_i) = \max - \frac{\max - \min}{2n(x)} - \frac{\max - \min}{2n(y)} \quad (23)$$

Where *max* and *min* are constants 2, and 1 respectively; n(x) and n(y) refer to the density of nodes x, y respectively, they are computed as follows:

$$n(c) = |neighbor(c)| \quad (24)$$

Where $neighbor(c)$ refer to the adjacent nodes of concept *c* in the taxonomic ontology, leaving c.

## 3.3 Comparison

For comparing the behavior of the above two path models in edge-based similarity measures, in this section, we chose the six popular edge-based algorithms to use the two path models respectively to measure similarity for word pairs in MC30 and RG65 [32] datasets, and then calculate the Pearson correlation coefficients between their measurements and human judgments. Table 1 summarizes the results of their correlation coefficients.

Table 1: Correlations of edge counting-based model and edge weight-based model on MC30 and RG65

| Path model | Measure | Formula | MC30 | RG65 |
|---|---|---|---|---|
| Edge counting-based | Rada et al. [11] | Eq. (1) | 0.6379 | 0.7374 |
| | Leacock et al. [19] | Eq. (2) | 0.7977 | 0.8518 |
| | Wu et al. [16] | Eq. (4) | 0.7464 | 0.7854 |
| | Liu_1 et al. [4] | Eq. (5) | 0.8018 | 0.8446 |
| | Liu_2 et al. [4] | Eq. (6) | 0.7711 | 0.8405 |
| | Li et al. [18] | Eq. (7) | 0.8008 | 0.8559 |
| Edge weight-based | Rada et al. [11] | Eq. (1) | 0.6448 | 0.7443 |
| | Leacock et al. [19] | Eq. (2) | 0.8038 | 0.8517 |
| | Wu et al. [16] | Eq. (4) | 0.7792 | 0.8213 |
| | Liu_1 et al. [4] | Eq. (5) | 0.8201 | 0.8579 |
| | Liu_2 et al. [4] | Eq. (6) | 0.8087 | 0.8684 |
| | Li et al. [18] | Eq. (7) | 0.8183 | 0.8595 |

From the results in Table 1, we observed that the edge weight based on density can slightly improve the similarity measures in Eq. (1), Eq. (2), Eq. (4), Eq. (5), Eq. (6) and Eq. (7), which are path-based approaches. Therefore, we can draw the conclusion that considering the density factor in path computation can improve the path-based measures, but the effect of using density-based weight to promote measuring accuracy is limited. This paper will propose a new density-based path computing model to greatly enhance the accuracy of edge-based similarity measures.

# 4. A new density-based path computing model

## 4.1 Proposed model

Our proposed model is derived from such a widespread phenomenon in similarity measures: In Fig.1, Suppose that the set of subsumers of concept $c_1$ or $c_1$ in their shortest path, including their Least Common Subsumer (LCS) and $S_i$, are represented by a function of LocalSubsumers ($c_1$,$c_2$), when the number of direct hyponyms of LocalSubsumers($c_1$,$c_2$) increases(from Fig. 1.(a) to Fig. 1.(b)), i.e. the density of LocalSubsumers($c_1$,$c_2$) becomes greater, the information content of LCS($c_1$,$c_2$) becomes smaller and the information content of concepts $c_1$ and $c_2$ are unchanged, so the similarity between $c_1$ and $c_2$ will decline in IC-based similarity measures according to the equation (12), (13) or equation (14). However, as the density of LocalSubsumers ($c_1$, $c_2$) becomes greater, due to the shortest path between concepts $c_1$and $c_2$ and the depth of their LCS are not changed, so the similarity between $c_1$ and $c_2$ will not change in edge-based similarity measures. This phenomenon reflects the main differences between the measures IC-based and edge-based. In order to make up the deficiencies of edge-based similarity measures in this phenomenon, we propose a path compensation model based on local area density with less computing time overhead.

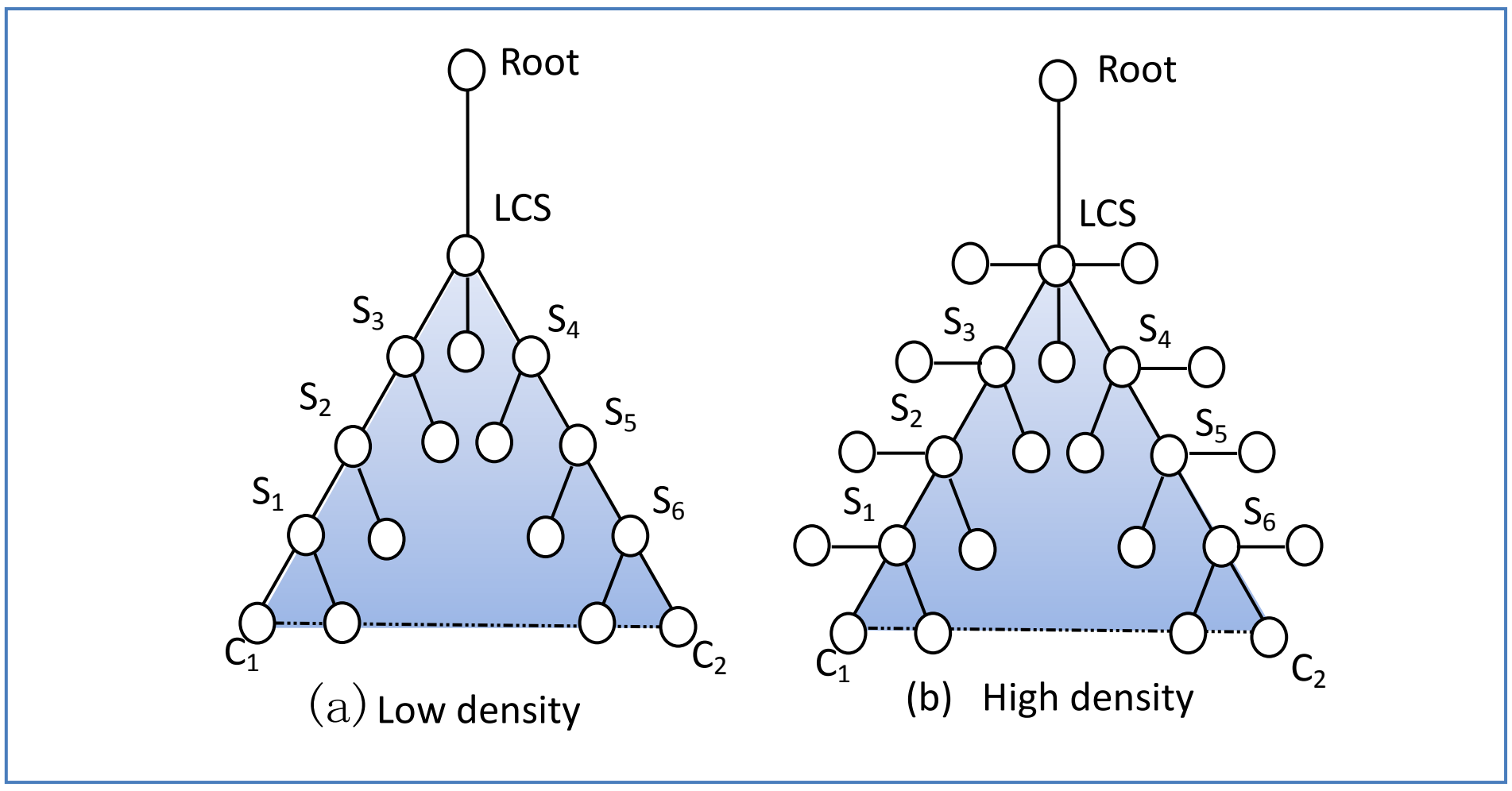

Fig. 1: An abstract diagram of the taxonomic ontology

**Definition 1.** Let C be the set of concepts in the taxonomic ontology, we define the local density of a concept as:

$$Density(c) = |\{h \in C \mid h \in Sons(c)\}| \tag{25}$$

Where, Sons(c) refer to the set of direct hyponyms of the concept c. Note that, the density of concept is only equal to the number of its direct hyponyms, which is directly linked by it, but not all the hyponyms.

**Definition 2.** Let the operation of subsumption $\leq$ be a binary relation: C×C, C refer to the set of concepts in the ontology. We define the set of the subsumers of concept $c_1$ or $c_2$ in their shortest path as the local area subsumers of concepts $c_1$ and $c_2$; and it is computed as follows:

$$AreaSubsumers(c_1,c_2) = \{s \in C \mid (c_1 \leq s \ \vee \ c_2 \leq s) \ \wedge \ s \text{ in path}(c_1,c_2)\} \tag{26}$$

Where the expression $c \leq s$ means that c is a hierarchical specialization of the concept s, path($c_1$,$c_2$) refer to the shortest path between concepts $c_1$ and $c_2$.

**Definition 3.** We define the local area density of concepts $c_1$ and $c_2$ as:

$$AreaDensity(c_1,c_2) = \sum_{s \in AreaSubsumers(c_1,c_2)} density(s) \tag{27}$$

**Definition 4.** We define the path compensation based on the local area density of concepts as:

$$PathComp(c_1,c_2) = \frac{AreaDensity(c_1,c_2)}{AreaDepth(c_1, LCS(c_1,c_2), c_2)} \times \lambda \tag{28}$$

Where, path ($c_1$,$c_2$) refer to the shortest path between the concepts $c_1$ and $c_2$, Edges (Path($c_1$, $c_2$)) to the set of edges in the path ($c_1$,$c_2$) , λ is an adjustable compensation factor and $\lambda \in [0,1]$, AreaDepth($c_1$,LCS($c_1$,$c_2$),$c_2$) refer to the average depth of the local area of the triangle △($c_1$,LCS($c_1$,$c_2$),$c_2$)(the Shaded part in Fig.1 ) which is the depth of the gravity center of that triangle , and it is computed as follows:

$$AreaDepth(c_1, LCS(c_1,c_2), c_2) = \frac{Depth(c_1) + Depth(LCS(c_1,c_2)) + Depth(c_2)}{3} \tag{29}$$

**Definition 5.** We define the path computing model based on the local area density of concepts as:

$$\begin{aligned} Path_{density}(c_1,c_2) &= Path_{edge}(c_1,c_2) + PathComp(c_1,c_2) \\ &= |\text{Edges}(\text{Path}(c_1,c_2))| + \frac{\text{AreaDensity}(c_1,c_2)}{\text{AreaDepth}(c_1,\text{LCS}(c_1,c_2),c_2)} \times \lambda \end{aligned} \tag{30}$$

In Eq. (27), we consider the local area density of concepts as an extension of the edge-based path, and convert the local area density divided by their depth into the compensation for edge-based path with an adjustable parameter, which is based on two such axioms: the semantic distance between concepts monotonically increases according to their local area density, and monotonically decreases according to the average depth of the local area where the concepts lie. Here we use the information theory to prove them, as shown in Table 2 and Table 3.

**Proposition 1.** ∀a,b,x,y∈C| IC(a)= IC(x) ∧ IC(b)=IC(y) ∧ AreaDensity(a,b)> AreaDensity (x,y) => Distance(a,b) > Distance(x,y).

**Table 2: Proof for Proposition 1**

| Step | Reasoning | Explanation |
|---|---|---|
| 1 | AreaDensity (a,b) > AreaDensity (x,y) ∧ IC(a)= IC(x) ∧ IC(b)=IC(y) => | As the local area density of concepts becomes larger whereas their IC are unchanged |
| 2 | \|Hypos(LCS(a, b)) \| > \|Hypos(LCS(x, y))\| => | the total of hyponyms of their LCS monotonically increases |
| 3 | IC(LCS(a, b)) < IC(LCS(x, y)) => | the IC of their LCS monotonically decreases |
| 4 | Sim(a,b) < Sim(x,y) => | their similarity monotonically decreases according to the equation (12), (13) or (14) |
| 5 | Distance(a,b) > Distance(x,y) | their semantic distance monotonically increases |

**Proposition 2.** ∀a,b,x,y∈C| IC(a)= IC(x) ∧ IC(b)=IC(y) ∧ AreaDensity(a,b)=AreaDensity (x,y) ∧ AreaDepth( a ,LCS(a, b), b) > AreaDepth( x, LCS(x, y),y) => Distance(a,b) < Distance(x,y).

**Table 3: Proof for Proposition 2**

| Step | Reasoning | Explanation |
|---|---|---|
| 1 | AreaDepth( a ,LCS(a, b), b) > AreaDepth( x, LCS(x, y),y) ∧AreaDensity (a,b) = AreaDensity (x,y) ∧ IC(a)= IC(x) ∧ IC(b)=IC(y) => | As the average depth of concepts' local area deepens whereas their IC and local area density are unchanged |
| 2 | $\sum_{c_1 \in Subsumers(a,b)} IC(c_1) > \sum_{c_2 \in AreaSubsumers(x,y)} IC(c_2)$ => | IC values monotonically increase as one moves down in the taxonomy [28] |
| 3 | $\sum_{s_1 \in Subsumers(a,b)} \|Hypos(s_1)\| < \sum_{s_2 \in Subsumers(x,y)} \|Hypos(s_2)\|$ => | the total of hyponyms of these subsumers monotonically decreases |
| 4 | IC(LCS(a, b)) > IC(LCS(x, y)) => | the IC of their LCS monotonically increases |
| 5 | Sim(a,b) > Sim(x,y) => | their similarity monotonically increases according to the equation (12), (13) or (14) |
| 6 | Distance(a,b) < Distance(x,y) | their semantic distance monotonically decreases |

## 4.2 Application in edge-based measures

Our proposed method is a generic path computing model and can be applied on various edge-based similarity algorithms. The specific steps are: the original structure of the algorithm formulas remains unchanged, we just use our path model to replace the edge counting-based path computation in the algorithm formulas, and the depth still is computed according to edge counting.

In Eq. (30), the adjustable compensation factor λ is an important parameter, which determines the scale in which density converts path length. The λ is an empirical value and is related to the training sets, taxonomic ontology and specific similarity algorithms. We chose six edge-based similarity algorithms to combine with our path computing model and measure the MC30 and RG65 datasets with different values of the parameter λ. Fig. 2 demonstrates how the Pearson correlation coefficients between computer-based measurements and human judgment are changed with the adjustable compensation factor λ. From Fig. 2, we observed that as λ=0.3, the Pearson correlation coefficients are at or near maximum in six selected similarity algorithms (as shown in Table 4). Though the initial correlations in different edge-based algorithms (λ=0) is not the same, but most of their best correlations on MC30 meet or exceed 0.85 by taking different values of the parameter λ for their path compensations(as shown in Table 5). This shows that our path model has the specific function to remedy the structural differences in edge-based similarity algorithms.

In summary, we can use the following two ways to apply our model:

(1) General way. When the parameter λ is equal to 0.3 in Eq. (30), six edge-based algorithms combined with our model have achieved better correlations on the MC30 and RG65 datasets, as shown in Table 4. In order to fit all solution, we take λ= 0.3 as a general way of our model.

(2) Best way. In order to obtain the best correlations of each method on MC30 or RG65 dataset, we can take the different values of parameter λ for different edge-based algorithm, as shown in Table 5.

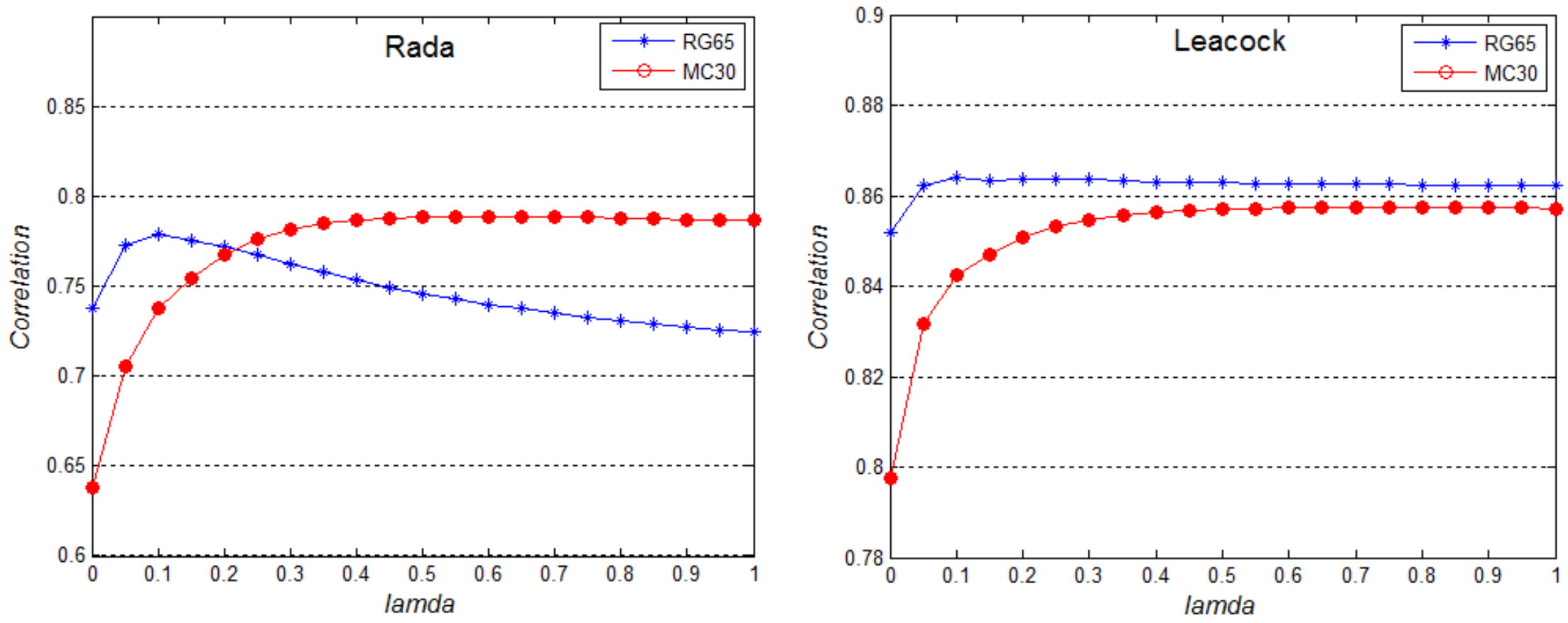

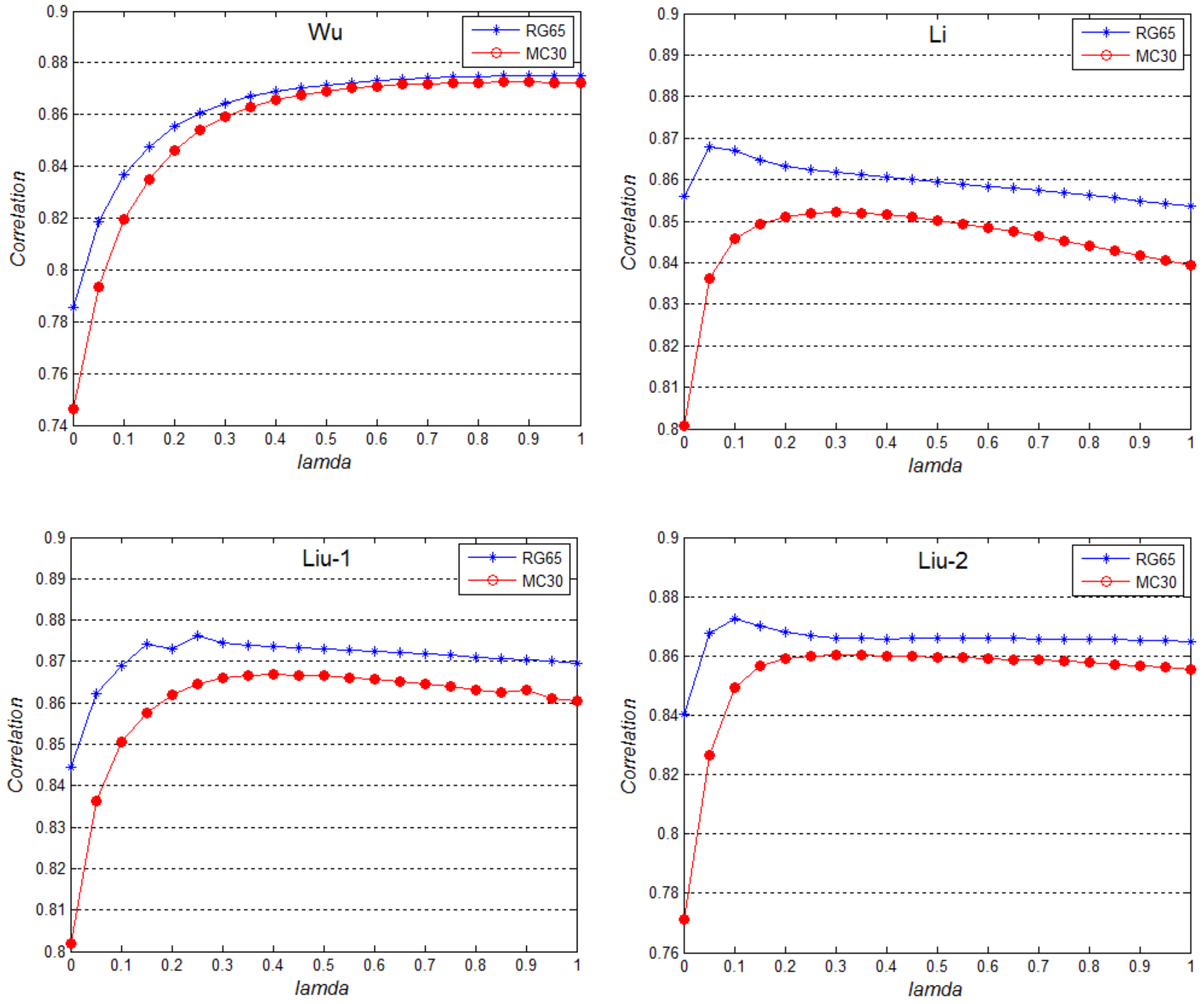

Fig. 2: Variation of the correlation coefficients of six edge-based similarity algorithms according to the compensation factor λ

Table 4 Better correlations of six edge-based algorithms combined with our general model on MC30 and RG65 datasets (λ=0.3)

| | Rada | Leacock | Wu | Liu-1 | Liu-2 | Li |
|---|---|---|---|---|---|---|
| Better correlation on MC30 | 0.7814 | 0.8545 | 0.8592 | 0.8659 | 0.8602 | 0.8522 |
| Better correlation on RG65 | 0.7623 | 0.8635 | 0.8642 | 0.8744 | 0.8661 | 0.8617 |

Table 5 Best correlations of each method with our model on MC30 and RG65 datasets

| | Rada | Leacock | Wu | Liu-1 | Liu-2 | Li |
|---|---|---|---|---|---|---|
| Best correlation on MC30 | 0.7888 (λ=0.6) | 0.8575 (λ=0.8) | 0.8725 (λ=0.9) | 0.8668 (λ=0.4) | 0.8602 (λ=0.3) | 0.8522 (λ=0.3) |
| Best correlation on RG65 | 0.7784 (λ=0.1) | 0.8639 (λ=0.1) | 0.8752 (λ=1.0) | 0.8762 (λ=0.3) | 0.8725 (λ=0.1) | 0.8669 (λ=0.1) |

# 5. Evaluation

## 5.1 Performance

In this study, we have evaluated the performance of the proposed path computation method from two aspects: First, we compared it with the edge counting-based path computing model through using the same edge-based similarity algorithms combined with different path models respectively to measure the same data set, from which to observe the ability of our model to enhance the measurement accuracy of the edge-based similarity algorithms; then we compared the edge-based similarity algorithms combined with our path model with the current various excellent similarity algorithms, including IC-based approaches and hybrid approaches, to evaluate whether the edge-based similarity algorithms combined with our path model has reached a level of excellence. In order to ensure fairness, we used the famous Miller & Charles [29] and Rubenstein & Goodenough [32] benchmark, which has become a defacto standard to evaluate the performance of similarity measures and many related works have taken this metric as test bed [12,17,22,27,30,31].

The dataset in Miller and Charles metric consists of 30 English noun pairs extracted from the original 65 in Rubenstein and Goodenough metric [32], and the similarity of each pair have been judged on a scale from 0 (semantically unrelated) to 4 (highly synonymous) by 38 participants. Semantic similarity measures can be evaluated using the Pearson correlation coefficients to correlate the scores computed by a measure with the judgments provided by humans in the MC30 dataset. In comparison experiments, we exploited WordNet 3.0 as the taxonomic ontology and adopt the JWI (Java WordNet Interface) [33] to query relate data in WordNet3.0 database, which was presided and wrote by Mark Alan.Finlayson who was from Massachusetts Institute of Technology Computer Science and Artificial Intelligence Lab.

Table 6 shows the correlations of six edge-based measures combined with different path models respectively, three IC-based measures combined with different IC computations respectively, three feature-based measures and two hybrid measures on MC30 and RG65 datasets. Table 7 shows the similarity scores for each word pair of MC30 in several similarity measures.

Table 6: Pearson correlation coefficients between different measures and human judgment on MC30 and RG65 datasets

| Similarity measure | Proposed in | Type | MC30 | RG65 | Evaluated in |
|---|---|---|---|---|---|
| Rada (path computed as edge counting) | [11] | Only path | 0.64 | 0.74 | This study |
| Leacock (path computed as edge counting) | [19] | Only path | 0.80 | 0.85 | This study |
| Wu (path computed as edge counting) | [16] | Depth and path | 0.75 | 0.79 | This study |
| Liu_1 (path computed as edge counting) | [4] | Depth and path | 0.80 | 0.84 | This study |
| Liu_2 (path computed as edge counting) | [4] | Depth and path | 0.77 | 0.84 | This study |
| Li (path computed as edge counting) | [18] | Depth and path | 0.80 | 0.86 | This study |
| Rada (path computed as our general model in | This study | Only path | 0.78 | 0.76 | This study |
| Eq.(30)( λ=0.3)) | | | | | |
| Leacock (path computed as our general model in Eq.(30)( λ=0.3)) | This study | Only path | 0.85 | 0.86 | This study |
| Wu (path computed as our general model in Eq.(30)( λ=0.3)) | This study | Depth and path | 0.86 | 0.86 | This study |
| Liu_1 (path computed as our general model in Eq.(30)( λ=0.3)) | This study | Depth and path | 0.87 | 0.87 | This study |
| Liu_2 (path computed as our general model in Eq.(30)( λ=0.3)) | This study | Depth and path | 0.86 | 0.87 | This study |
| Li (path computed as our general model in Eq.(30)( λ=0.3)) | This study | Depth and path | 0.85 | 0.86 | This study |
| Resnik (IC computed as Seco et al.) | [27] | IC (intrinsic) | 0.77 | 0.85 | [27],[10] |
| Lin (IC computed as Seco et al.) | [27] | IC (intrinsic) | 0.81 | 0.82 | [27],[10] |
| Jiang (IC computed as Seco et al.) | [27] | IC (intrinsic) | 0.84 | 0.85 | [27] ,[10] |
| Resnik (IC computed as Sanchez et al.) | [28] | IC (intrinsic) | 0.84 | 0.86 | [28] ,[10] |
| Lin (IC computed as Sanchez et al.) | [28] | IC (intrinsic) | 0.85 | 0.85 | [28] ,[10] |
| Jiang (IC computed as Sanchez et al.) | [28] | IC (intrinsic) | 0.87 | 0.87 | [28] ,[10] |
| Rodriguez and Egenhofer | [34] | Feature | 0.71 | 0.78 | [10] |
| Petrakis et al. | [31] | Feature | 0.68 | 0.78 | [10] |
| David et al. | [36] | Feature | 0.83 | 0.86 | [36] |
| Mohamed | [10] | Hybrid | 0.85 | 0.88 | [10] |
| Zhou | [35] | Hybrid | 0.86 | 0.87 | [10] |

Table 7: Similarity scores for each word pair of MC30 in several similarity measures

| Word 1 | Word 2 | MC30 (normalization) | Wu( path computed as edge counting) | Wu (path computed as in Eq. (30) λ=0.3) | Liu-2( path computed as edge counting) | Liu-2 (path computed as in Eq. (30) λ=0.3) |
|---|---|---|---|---|---|---|
| car | automobile | 0.98 | 1.0 | 1.0 | 1.0 | 1.0 |
| gem | jewel | 0.96 | 1.0 | 1.0 | 1.0 | 1.0 |
| journey | voyage | 0.96 | 0.9474 | 0.9474 | 0.9676 | 0.9676 |
| boy | lad | 0.94 | 0.9412 | 0.9412 | 0.9574 | 0.9574 |
| coast | shore | 0.925 | 0.8889 | 0.8688 | 0.8581 | 0.8298 |
| asylum | madhouse | 0.9025 | 0.9474 | 0.9474 | 0.9676 | 0.9676 |
| magician | wizard | 0.875 | 1.0 | 1.0 | 1.0 | 1.0 |
| midday | noon | 0.855 | 1.0 | 1.0 | 1.0 | 1.0 |
| furnace | stove | 0.7775 | 0.4706 | 0.2862 | 0.1684 | 0.0117 |
| food | fruit | 0.77 | 0.3077 | 0.1974 | 0.0710 | 0.0112 |
| bird | cock | 0.7625 | 0.9474 | 0.9474 | 0.9676 | 0.9676 |
| bird | crane | 0.7425 | 0.8571 | 0.8299 | 0.8837 | 0.8485 |
| tool | implement | 0.7375 | 0.9231 | 0.8321 | 0.9246 | 0.8072 |
| brother | monk | 0.705 | 0.9474 | 0.9426 | 0.9676 | 0.9642 |
| crane | implement | 0.42 | 0.7143 | 0.4792 | 0.5917 | 0.1498 |
| lad | brother | 0.415 | 0.7500 | 0.3590 | 0.6696 | 0.0162 |
| journey | car | 0.29 | 0.0 | 0.0 | 0.0 | 0.0 |
| monk | oracle | 0.275 | 0.6316 | 0.3324 | 0.4227 | 0.0084 |
| cemetery | woodland | 0.2375 | 0.3333 | 0.1702 | 0.0922 | 0.0050 |
| food | rooster | 0.2225 | 0.1176 | 0.0691 | 0.0068 | 0.0003 |
| coast | hill | 0.2175 | 0.6000 | 0.4776 | 0.3940 | 0.2118 |
| forest | graveyard | 0.21 | 0.3333 | 0.1702 | 0.0922 | 0.0050 |
| shore | woodland | 0.1575 | 0.5000 | 0.2601 | 0.2741 | 0.0385 |
| monk | slave | 0.1375 | 0.7500 | 0.3529 | 0.6696 | 0.0141 |
| coast | forest | 0.105 | 0.4444 | 0.2507 | 0.2067 | 0.0333 |
| lad | wizard | 0.105 | 0.7500 | 0.3630 | 0.6696 | 0.0178 |
| chord | smile | 0.0325 | 0.2857 | 0.1917 | 0.0548 | 0.0096 |
| glass | magician | 0.0275 | 0.4000 | 0.1462 | 0.1163 | 0.0002 |
| noon | string | 0.02 | 0.1538 | 0.0910 | 0.0190 | 0.0019 |
| rooster | voyage | 0.02 | 0.0 | 0.0 | 0.0 | 0.0 |
| Correlation with MC30 | | - | 0.7464 | 0.8592 | 0.7711 | 0.8602 |

## 5.2 Efficiency

Efficiency is an important indicator to assess the usefulness of a method. IC-based similarity measures require to count the number of all hyponyms of the concepts in the taxonomy it average time complexity is O(Max_Nodes) in theory (in WordNet3.0, the Max_Nodes is equal to 82115); while edge-based similarity measures only needs to search the branch where the concepts lie, its average time complexity is O(Max_Depth) in theory (in WordNet3.0, the Max_Depth is equal to 19).In order to validate the big differences between the edge-based measures combined with our path model and IC-based measures in time complexity, we selected the Wu & Palmer’s method (edge-based) and the Lin’ s method (IC-based) for comparing in terms of efficiency. Choosing these two methods for comparison was due to the similar formula structures of these two methods, their similarities are equal to the double commonality between two concepts divided by their complete descriptions or complete feature. After testing, most of the time complexities of the edge-based measures are in the same order of magnitude, the same is true on the IC-based measures. Computer configuration used in our experiment is shown in Table 8. The experiment results are shown in Table 9. The column *Totaltime* in Table 9 refers to the total time for benchmark, *AverageTime* to the average time for each word pair, and their units are seconds.

As stated in Section 1, the development trend of taxonomic ontology is online and real-time updates. In order to accommodate this trend, we assume that WordNet is a real-time dynamic ontology, rather than experience. So, in IC-based similarity measures, we use the following formula to calculate the total time for each measurement:

$$TotalTime = PretreatmentTime + ComputingTime \quad (31)$$

Where, given that the subsumption relationship is recursive, the *PretreatmentTime* is used to explore the set of hyponyms of the root node to perfectly characterize the rest of concepts that, obviously, are specializations of the root, and count and store the number of all hyponyms of all concepts in a hash table. The *ComputingTime* is used for the IC-based algorithms to compute the similarity scores of each word pair on MC30 or RG65 according to the hash table of hyponym.

Table 8: Computer configuration used in the experiment

| Computer type | CPU type | CPU frequency | Memory |
|---|---|---|---|
| Desktop PC | i5-2400 | 3.1GHz | 4GB |

Table 9: Efficiency Comparison between the edge-based and IC-based measures (Units: seconds)

| | Pretreatment | Computation | TotalTime | AverageTime | Dataset |
|---|---|---|---|---|---|
| edge-based measures (path computed as edge counting) | 0 | 2.27 | 2.27 | 0.08 | On MC30 |
| | | 4.56 | 4.56 | 0.07 | On RG65 |
| edge-based measures (path computed as in Eq. (30)) | 0 | 3.68 | 3.68 | 0.12 | On MC30 |
| | | 5.71 | 5.71 | 0.09 | On RG65 |
| IC-based measures | 159.57 | 4.42 | 163.99 | 5.47 | On MC30 |
| | | 6.02 | 165.59 | 2.55 | On RG65 |

# 6. Discussion

From the above experiment results, we can draw several conclusions. First, the results form Table 4 to Table 7 show our path model can greatly improve the measurement accuracy of various edge-based similarity algorithms, including the path-based and Path & depth –based algorithms. Combined with our path model, there are five edge-based algorithms to obtain the correlation of over 0.85 and the best correlation reaches 0.87 on MC30, which is the widely recognized repeatable highest level [10,35] of computer-based similarity measures on MC30 dataset and quite close the average correlation (0.9015) between individual subjects of human replication reported in Resnik's replication [12] of Miller and Charles experiment. Through the analysis, we found that good result mainly thanks to the capacity that our path model effectively reduce the impact of high local area density on similarity measurement, for example, for the word pairs monk & slave, coast & forest, lad & wizard, etc., converting their local area densities into their path by our path model greatly improved their similarity accuracy, as shown in Table 7. Therefore, our model successfully solved the non-uniformity problem of taxonomic ontology.

With regards to the equivalence with IC-based measures, from the results in Table 6, it is observed that the role of our path model in the edge-based measures is equivalent with the depth in the IC-based measures. Before the introduction of depth, the results in IC-based measures(IC computed as Seco et al.) are not satisfactory; and the introduction of the depth of subsumers (IC computed as Sanchez et al.) improves the average correlation of IC-based measures on MC30 from 0.8 to more than 0.85. Similarly, our path model improves the average correlation of edge-based measures on MC30 from less than 0.8 to more than 0.85.

With regards to the measuring efficiency, the results in Table 9 confirmed that our path model has a big advantage in efficiency than IC computation. The edge-based measures combined with our path model are almost the same faster as the measures based on edge counting, and more than several dozen times faster than IC-based measures. This means that the edge-based measures combined with our model has a better application prospect than IC-based measures. In fact, our model is essentially a way to combine the edge-based path with density skillfully. The density computation in our model only need to count the number of direct hyponyms of concept, rather than all the hyponyms, so it can save a lot of computing time. In contrast, the IC computation must completely count the number of all the hyponyms or leaves of concept, so it takes a lot of computing time. Although the efficiency of IC computation can be improved through the precomputing the number of all hyponyms of all the concepts and saving them in a file, but this method is powerless for a real-time updated taxonomic ontology.

# 7. Conclusions

Edge is an important component of the hierarchical structure of a taxonomic ontology. The edge-based similarity measure has the advantages of simple, intuitive and easy to understand and it has better performance in some specific applications of computational intelligence with highly constrained taxonomies. But edge-based similarity measures are encountering the problems of the non-uniformity in a large taxonomic ontology such as WordNet. Though, this problem can be better corrected by the IC-based measure, but the IC computation has to consume a lot of computing time in a dynamic ontology, this is difficult to bear in some real-time applications such as QA systems and Web information retrieval. The proposed model broke this deadlock; it can greatly improve the measurement accuracy of various edge-based similarity algorithms with less computing time overhead. Experiments show that our model has broad application prospects.

Our model is essentially a way to combine the edge-based path with density skillfully. It considers the local area density of concepts as an extension of the edge-based path and converts the local area density divided by their depth into the compensation for edge-based path with less computing time overhead, which successfully solved the non-uniformity problem of taxonomic ontology. Through this study, we believe more firmly that the similarity measure in a large taxonomic ontology require to combine multi-source information extracted from the ontology. In future, we will work to find out more semantic evidences in WordNet and integrate them with the edge-based similarity measure to challenge the theoretical upper bound (0.9015) of the correlation between computer-based measurement and human judgment.